\documentclass[11pt]{article}
\pdfoutput=1
\usepackage[margin=1in]{geometry}
\usepackage{xcolor}
\usepackage[utf8]{inputenc}
\usepackage[T1]{fontenc}
\usepackage{hyperref}
\usepackage{url}
\usepackage{booktabs}
\usepackage{amsfonts}
\usepackage{nicefrac}
\usepackage[protrusion=true,expansion=false]{microtype}
\usepackage{amsmath}
\usepackage{graphicx}
\usepackage{subcaption}
\usepackage{multirow}

\graphicspath{{media/}}

\title{Stable FP4 Training via Transposition-Invariant Block Quantization}

\author{
	Mehdi Rahimifar \and
	Amin Darabi \and
	Xing Huang \and
	Zhijun Tu \and
	Yunke Peng \and
	Mehran Taghian Jazi \and
	Yao Wang \and
	Yufei Cui \and
	Hongliang Li
}
\date{}
\begin{document}
	
	\maketitle

\begin{abstract}

Reducing training precision is a key lever for improving the efficiency of large language model (LLM) training, but pushing beyond FP8 to 4-bit floating point (FP4) remains challenging due to instability during optimization. We identify a fundamental source of this instability in existing microscaling approaches: \emph{scale inconsistency induced by tensor transposition}. In conventional 1D block quantization, forward and backward passes assign different scaling factors to the same values after transposition, leading to biased and unstable gradient updates. To address this issue, we propose a low-precision training framework based on \emph{2D block FP4 quantization}, which enforces transposition-invariant scaling and preserves consistency between forward and backward computations. We further combine this with truncation-free scaling and stochastic rounding to control quantization error and maintain unbiased gradients. To handle the sensitivity of attention mechanisms, we adopt MXFP8 quantization for query and key projections, yielding a practical mixed-precision design. We evaluate our method on dense LLMs up to 7B parameters and a 30B Mixture-of-Experts model, trained on up to 100B tokens. Across all settings, our approach achieves stable end-to-end FP4 training and closely matches BF16 performance, with less than 1.3\% degradation in perplexity and downstream accuracy. These results demonstrate that enforcing forward--backward scaling consistency is sufficient to enable practical FP4 training at scale, providing a simple and effective pathway toward more efficient LLM training.

\end{abstract}

\section{Introduction}

Large Language Models (LLMs) have enabled major advances in natural language processing, including reasoning, code generation, and multimodal understanding~\cite{agarwal2025gpt,gong2025survey,minaee2025largelanguagemodelssurvey}. However, these gains come at rapidly increasing computational and energy costs, making efficient training a central challenge in scaling LLMs~\cite{lang2024comprehensive,jegham2025hungryaibenchmarkingenergy}. Reducing numerical precision has emerged as a key approach to improving efficiency, as lower-precision formats can significantly reduce memory bandwidth and increase arithmetic throughput~\cite{sun2025scaling,hernandez2025towards,hao2025low}. While FP8 training is now practical on modern accelerators, pushing to 4-bit floating point (FP4) promises further reductions in memory footprint and compute \cite{zhou2025towards,mishra2025recipes}. However, stable end-to-end FP4 training remains an open challenge due to limited dynamic range, large quantization error, and sensitivity during backpropagation and attention computation \cite{castro2025quartet,panferov2026quartet}.


\begin{figure}[t]
    \centering
    \includegraphics[width=0.9\linewidth]{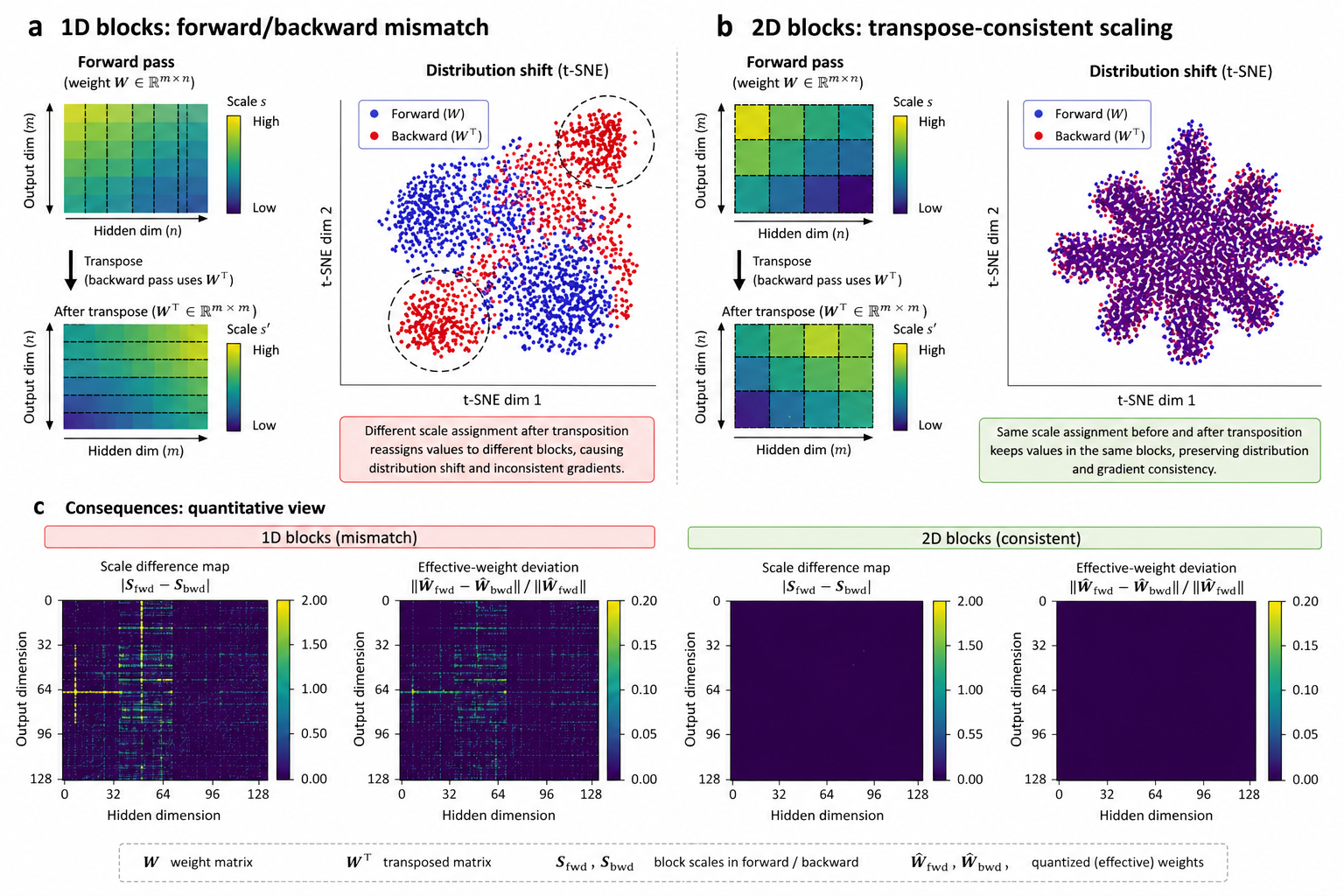}
   \caption{
\textbf{Forward--backward inconsistency in 1D microscaling  vs 2D blocks.}
(a) In 1D block quantization, transposition reassigns values to different scaling blocks, causing forward--backward mismatch and distribution shift.
(b) 2D block quantization preserves block structure under transposition, yielding consistent scaling and aligned distributions.
(c) Scale difference maps show large mismatch in 1D but near-zero error in 2D.
(d) Effective weight deviation is high for 1D blocks and negligible for 2D, confirming improved consistency.
}
    \label{fig:precisions}
\end{figure}

Recent microscaling approaches, such as MXFP4 and NVFP4, address these challenges using block-wise quantization, where groups of values share a scaling factor~\cite{yang2025empirical,tseng2025training,abecassis2025pretraining}. While effective in reducing quantization error, these methods share a critical limitation: they rely on \emph{one-dimensional (1D) block structures} that are not invariant to tensor transposition. During backpropagation, weight and activation matrices are transposed, reassigning values to different scaling blocks. This induces \emph{scale inconsistency} between forward and backward passes, leading to biased gradients and degraded training stability as shown in Fig \ref{fig:precisions}. \textbf{In this work, we identify transpose-induced scale inconsistency as a key failure mode in FP4 training.}
We show that 1D microscaling assigns inconsistent quantization scales across forward and backward passes, leading to biased gradients and unstable optimization. To address this, we propose a transposition-invariant FP4 training framework based on 2D block quantization, which enforces consistent scaling. We further combine this with truncation-free scaling and stochastic rounding, and adopt MXFP8 for query and key projections to stabilize attention. This design enables stable end-to-end FP4 training across models up to 30B parameters.

\paragraph{Contributions.}
\begin{itemize}
\item \textbf{Scale inconsistency as a failure mode.} We identify transposition-induced scale mismatch in 1D microscaling as a key source of instability in FP4 training.

\item \textbf{Transposition-invariant FP4 quantization.} We propose 2D block FP4 with truncation-free scaling and stochastic rounding to ensure consistent forward--backward behavior.

\item \textbf{Practical mixed-precision FP4 training.} We combine FP4 linear layers with MXFP8 attention and demonstrate stable training up to 30B models with $\sim$1\% gap to BF16.
\end{itemize}

\section{Motivation for Microscaling}

Efficient training of LLMs requires balancing numerical precision, dynamic range, and hardware efficiency. While reduced-precision formats such as FP16 and BF16 provide sufficient dynamic range, they remain costly in terms of memory bandwidth and compute, which increasingly dominate training at scale. Microscaling (MX) formats address this limitation by combining low-bit floating-point representations with block-wise scaling~\cite{rouhani2023microscaling}. Instead of using a single global scale, MXFP assigns shared scaling factors to small groups of values, preserving local dynamic range while enabling aggressive precision reduction~\cite{mishra2025recipes,hernandez2025towards}. Among these, MXFP4 is particularly attractive due to its extreme compression, using an \texttt{E2M1} representation with only four bits per value~\cite{wang2025optimizing}. However, this aggressive quantization amplifies sensitivity to scaling errors, making stable training challenging.

\paragraph{Key limitation.}
Existing microscaling approaches primarily reduce quantization error \emph{within} blocks, but largely overlook consistency \emph{across} forward and backward passes. As we show below, this inconsistency becomes a dominant source of instability in FP4 training.

\subsection{MXFP4 }

MXFP4 performs quantization using block-wise scaling:
\[
P_i = \text{round}_{\text{FP4}}\!\left( \frac{X_i}{S} \right), 
\qquad
X_i \approx P_i \cdot S,
\]
where $S$ is chosen based on the block maximum. NVFP4 improves representation fidelity by using smaller blocks and higher-precision scaling factors~\cite{abecassis2025pretraining}. While this reduces intra-block quantization error, both approaches share a fundamental structural limitation.

\paragraph{Scale inconsistency under transposition.}
Both methods use one-dimensional (1D) block layouts (e.g., $1\times32$ or $1\times16$). During backpropagation, tensors are transposed, which reassigns values to different blocks and therefore different scaling factors. This induces a mismatch between forward and backward quantization:
\[
\text{quant}_{\text{fwd}}(X) \neq \text{quant}_{\text{bwd}}(X),
\]
leading to biased gradients and unstable optimization (Fig.~\ref{fig:precisions}). Reducing block size mitigates this effect but does not eliminate it \cite{chmiel2025fp4}, and introduces additional overhead. This reveals a fundamental requirement: \emph{quantization schemes must be invariant to tensor transposition to ensure stable training}. 

\paragraph{Bias induced by scale inconsistency.}
To understand this instability, consider a tensor $X$ quantized in the forward pass with scale $S_{\text{fwd}}$ and in the backward pass with a different scale $S_{\text{bwd}}$. The resulting quantized values are
\[
\hat{X}_{\text{fwd}} = \mathrm{round}\!\left(\frac{X}{S_{\text{fwd}}}\right) S_{\text{fwd}}, 
\quad
\hat{X}_{\text{bwd}} = \mathrm{round}\!\left(\frac{X}{S_{\text{bwd}}}\right) S_{\text{bwd}}.
\]
When $S_{\text{fwd}} \neq S_{\text{bwd}}$, the forward and backward passes operate on different quantized representations. As a result, the backward pass effectively computes gradients with respect to a \emph{different quantized objective} than the one used in the forward pass. 

Under the straight-through estimator (STE) \cite{bengio2013estimating}, gradients are propagated as if $\hat{X} \approx X$, but this mismatch introduces a systematic bias:
\[
\nabla \mathcal{L}_{\text{bwd}} \;\neq\; \nabla \mathcal{L}_{\text{fwd}}.
\]
In contrast, if the quantization scheme is invariant to transposition (i.e., $S_{\text{fwd}} = S_{\text{bwd}}$), both passes operate on a consistent representation, eliminating this source of bias.

\paragraph{Implication.}
These observations suggest that stable FP4 training requires not only low quantization error, but also \emph{forward--backward consistency}. This motivates the design of transposition-invariant quantization schemes, which we achieve using 2D block structures in the next section.

\section{ Transposition-Invariant Block Quantization}

\begin{figure}[t]
    \centering
    \includegraphics[width=0.85\linewidth]{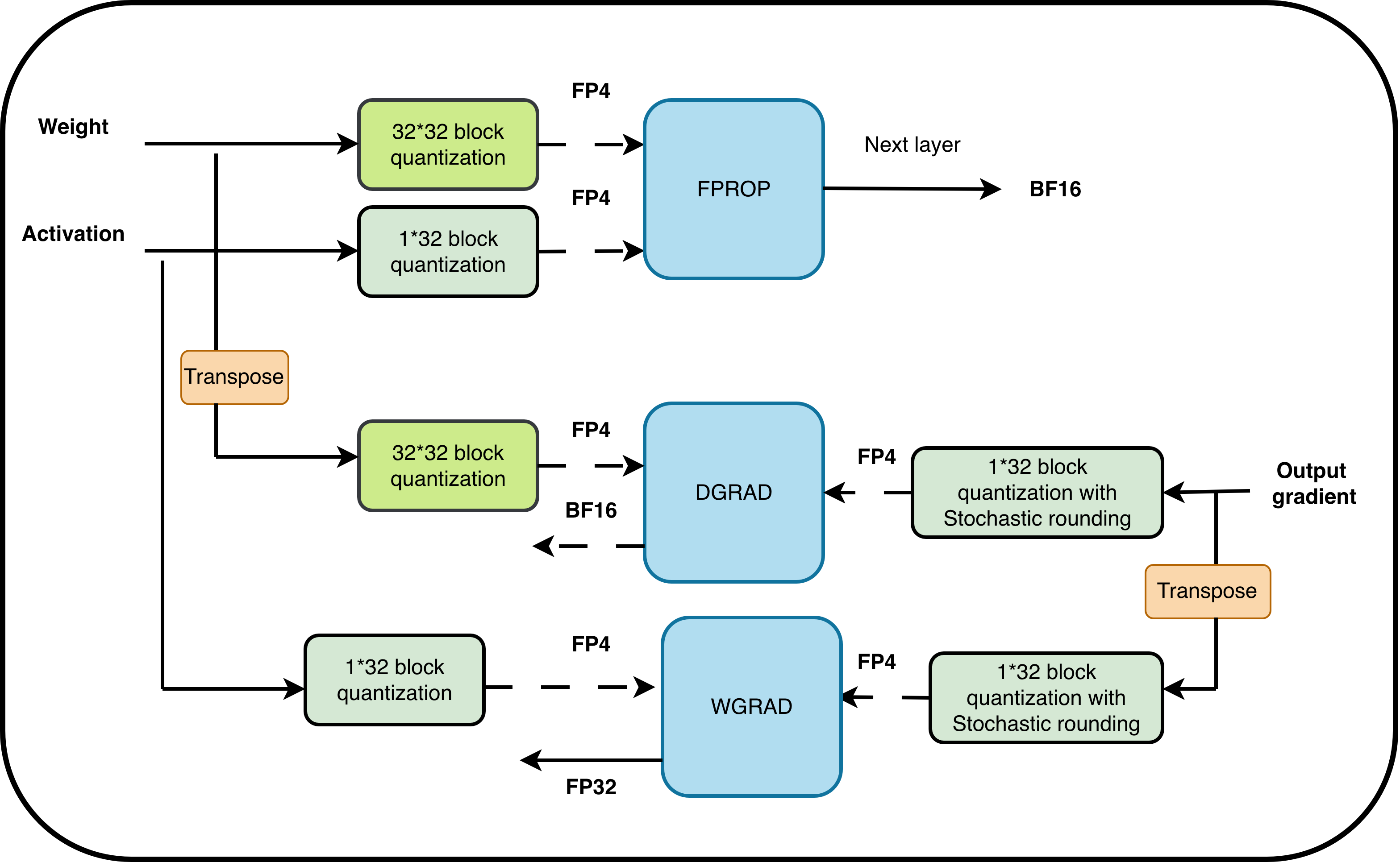}
    \caption{\textbf{Overview of 2D block FP4 quantization.} Tensors are partitioned into square blocks that share a scale, enabling consistent scaling across forward and backward passes.}
    \label{2D block FP4}
\end{figure}

We propose a transposition-invariant quantization scheme based on \emph{2D square blocks}. Unlike 1D layouts, square blocks preserve their structure under transposition, ensuring that values are associated with the same scaling factors in both forward and backward passes. The overall flow is illustrated in Fig \ref{2D block FP4}

\paragraph{Transposition invariance.}
Let $X \in \mathbb{R}^{m \times n}$ be partitioned into square blocks of size $b \times b$. A block $B_{i,j}$ in $X$ maps under transposition to the block $B_{j,i}$ in $X^\top$. Because $B_{j,i}$ contains the same values as $B_{i,j}$ up to transposition, the block maximum and therefore the block scale are preserved:
\[
S(B_{i,j}) = S(B_{i,j}^{\top}).
\]
This preserves the scale assignment of values across forward and backward computations. In contrast, 1D layouts generally regroup values after transposition, producing inconsistent quantization ranges.

\paragraph{Application to linear layers.}
For a linear layer:
\[
Y = X W^\top, \quad
\nabla X = \nabla Y W, \quad
\nabla W = (\nabla Y)^\top X,
\]
we apply 2D FP4 quantization to weights and gradients. We use $32\times32$ blocks for weights to align with hardware tiling. For gradients, we apply 2D quantization to both weight gradients and data gradients, enabling low-precision computation throughout the backward pass. We retain a 1D layout for activations to avoid mixing semantically unrelated values across channels. This reflects a trade-off between statistical structure and quantization consistency. To enable end-to-end training, we use the Straight-Through Estimator (STE)~\cite{bengio2013estimating}. Since quantization is piecewise constant and non-differentiable, directly computing gradients would block optimization. The STE approximates the derivative of the quantization function as identity during backpropagation. Concretely, for $\hat{X} = \mathrm{quant}(X)$, we use
\[
\frac{\partial \mathcal{L}}{\partial X}
\approx
\frac{\partial \mathcal{L}}{\partial \hat{X}}.
\]
This allows gradients to pass through the quantization step while retaining low-precision computation in the forward pass.

\subsection{Truncation-Free Scaling and Rounding}

The stability of low-precision training depends not only on block structure, but also on how values are scaled and rounded during quantization. In FP4, the limited dynamic range makes the system particularly sensitive to overflow and rounding bias, both of which can destabilize optimization if not properly controlled. To address overflow, we adopt truncation-free scaling. Specifically, we compute the scale as
\[
S = 2^{\,\lceil \log_2 (2M / (Q_p - Q_n)) \rceil},
\]
where $M$ is the maximum absolute value within a block and $(Q_p, Q_n)$ denote the positive and negative bounds of the FP4 format \cite{chen2025oscillation}. This choice ensures that all values fall within the representable range, preventing clipping and avoiding systematic distortion of large-magnitude activations and gradients. To mitigate bias introduced by quantization, we use stochastic rounding during backpropagation \cite{chmiel2025fp4}. Instead of deterministically mapping values to the nearest representable level, stochastic rounding preserves expectations:
\[
\mathbb{E}[\text{roundS}(x)] = x.
\]
This property is critical for maintaining unbiased gradient estimates, especially under aggressive quantization. Taken together, truncation-free scaling and stochastic rounding provide complementary benefits. The former eliminates instability caused by overflow, while the latter reduces bias in gradient estimation. In combination with transposition-invariant quantization, these mechanisms enable stable optimization under FP4 precision.

\subsection{MXFP8 Attention}

Attention computation is particularly sensitive to quantization due to two compounding effects: the dot-product interaction $QK^\top$ amplifies quantization noise, and the softmax normalization further magnifies small perturbations. As a result, aggressively reducing precision in this pathway often leads to instability or degraded model quality \cite{xi2024coat}. In practice, we observe that directly quantizing query ($Q$) and key ($K$) projections to FP4 introduces errors that propagate through $QK^\top$ and distort the attention distribution.
\begin{figure}[t]
    \centering
    \includegraphics[width=0.85\linewidth]{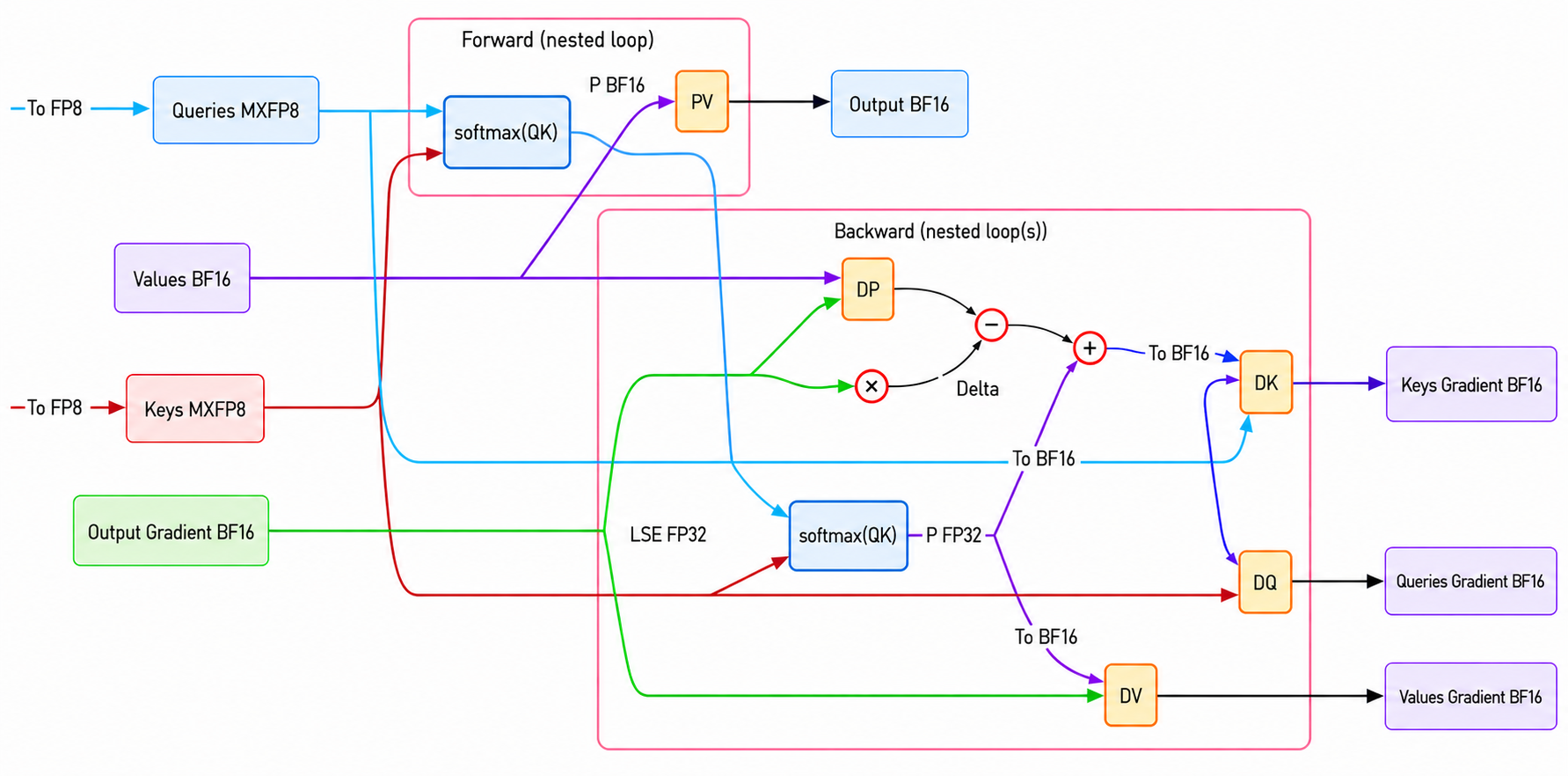}
    \caption{\textbf{MXFP8 attention pathway.} Query and key projections are quantized using MXFP8 to improve numerical stability while retaining low-precision computation in the rest of the model.}
    \label{fig:attention}
\end{figure}

While our 2D block FP4 quantization resolves forward--backward scale inconsistency in linear layers, it does not fully address this sensitivity in attention. To mitigate this issue, we adopt a mixed-precision design in which $Q$ and $K$ are quantized using MXFP8, while all other transformer linear layers (including MLP and attention value/output projections) use 2D block FP4. This preserves higher precision where it is most critical, while retaining the efficiency benefits of FP4 for the dominant matrix multiplications. The overall attention pipeline is illustrated in Figure~\ref{fig:attention}. Despite using higher precision for $Q$ and $K$, the overhead of this hybrid design remains modest. Empirically, we observe that the total training memory remains close to the FP4-only configuration, while providing improved numerical stability. 

To quantify the efficiency implications of this design, Table~\ref{tab:olmo1b_compact_efficiency} presents a theoretical comparison against BF16 training on OLMo-1B. We report weight memory and idealized throughput for transformer linear layers, which dominate training cost. Because the majority of parameters reside in MLP layers, which are fully quantized to FP4, the overall system retains most of the benefits of aggressive low-precision training. In this setting, our method achieves an estimated $\sim$65\% reduction in model weight memory and up to $\sim$3.6$\times$ idealized throughput improvement for linear operations, while only requiring FP8 precision in a small fraction ($\sim$12.5\%) of parameters.

\begin{table}[t]
\centering
\footnotesize
\caption{
Theoretical efficiency comparison on OLMo-1B.
Our method applies MXFP8 to Q/K linear layers and 2D-FP4 to all other transformer linear layers.
Memory values reflect parameter storage only; throughput estimates are idealized.
}
\label{tab:olmo1b_compact_efficiency}
\begin{tabular}{lcccc}
\toprule
Metric & BF16 & Our method & Saving / speedup & Notes \\
\midrule
Q/K linear weights & 268.4 MB & 134.2 MB & 50.0\% smaller & MXFP8 \\
Other attention linear weights & 268.4 MB & 67.1 MB & 75.0\% smaller & 2D-FP4 \\
MLP linear weights & 1610.6 MB & 402.7 MB & 75.0\% smaller & 2D-FP4 \\
\midrule
Transformer linear weights & 2147.4 MB & 604.0 MB & 71.9\% smaller & mixed \\
Total model weights & 2353.5 MB & 810.0 MB & 65.6\% smaller & embeddings BF16 \\
\midrule
Linear activation bandwidth & $1.00\times$ & $\sim0.30\times$ & $\sim70\%$ lower & weighted avg. \\
Ideal linear throughput & $1.00\times$ & $\sim3.56\times$ & $\sim3.56\times$ faster & ideal kernels \\
\bottomrule
\end{tabular}
\end{table}

Overall, combining 2D block FP4 quantization with MXFP8 attention yields a system that targets the two dominant costs in transformer training: matrix multiplications ($\mathcal{O}(Nd^2)$), which benefit from FP4, and attention computation ($\mathcal{O}(N^2 d)$), which benefits from higher precision in $QK^\top$. This balance enables stable end-to-end FP4 training while maintaining a favorable trade-off between efficiency and model quality.


\section{Experiments}

We evaluate whether enforcing forward--backward scaling consistency enables stable end-to-end FP4 training across model scales and architectures. Our experiments are designed to answer four questions:  
(1) Does 2D block FP4 eliminate the instability observed in conventional FP4 training?  
(2) How closely does the proposed method match BF16 training in convergence and final quality?  
(3) What is the contribution of truncation-free scaling, stochastic rounding, and MXFP8 attention?  
(4) How does the method behave across dense and mixture-of-experts models?

\subsection{Experimental Setup}

We consider three representative settings spanning both dense and sparse transformer architectures: OLMo-1B, OLMo-7B, and Qwen 30B MoE. These experiments allow us to assess whether the proposed method remains stable as model size and architectural complexity increase.

For each model, we compare the following configurations:
\begin{enumerate}
    \item \textbf{BF16}: a full-precision training baseline.
    \item \textbf{2D-FP4}: our proposed 2D block FP4 quantization applied to linear layers in both forward and backward passes.
    \item \textbf{2D-FP4 + MXFP8}: the same configuration augmented with MXFP8 quantization for attention queries and keys.
\end{enumerate}

For OLMo-1B, we train from scratch on a 100B-token corpus. We use the same token budget for OLMo-7B to evaluate scaling behavior in larger dense models, and for Qwen 30B MoE to assess robustness in a sparse mixture-of-experts setting.

Because native FP4 tensor-core support is not available on our training platform, we emulate microscaling behavior in software while training. Consequently, our experiments focus on \emph{optimization stability} and \emph{model quality}, rather than realized hardware speedup. 
Our method applies 2D block FP4 quantization consistently across forward and backward passes for linear layers. In the hybrid configuration, attention queries and keys are quantized using MXFP8, allowing us to isolate the benefit of higher precision in the most numerically sensitive part of the transformer. We evaluate models on both language modeling and downstream reasoning tasks. Language modeling performance is measured using perplexity on Wikitext, Pile, and C4, reported in Table~\ref{tab:lm_results}. Downstream reasoning performance is measured using accuracy on SciQ, COPA, ARC-Easy, and HellaSwag, reported in Table~\ref{tab:reasoning_results}. Training dynamics across model scales are shown in Figure~\ref{fig:main_results}, and ablations on quantization design choices are shown in Figure~\ref{fig:ablation_main}.

\subsection{Main Results}

We evaluate the proposed method along four dimensions: training stability, the effect of quantization design choices, language modeling performance, and downstream reasoning ability. Across all experiments, we compare BF16 training with our 2D block FP4 method, with and without MXFP8 attention.

\begin{figure}[t]
    \centering
    
    \begin{subfigure}[t]{0.55\linewidth}
        \centering
        \includegraphics[width=\linewidth]{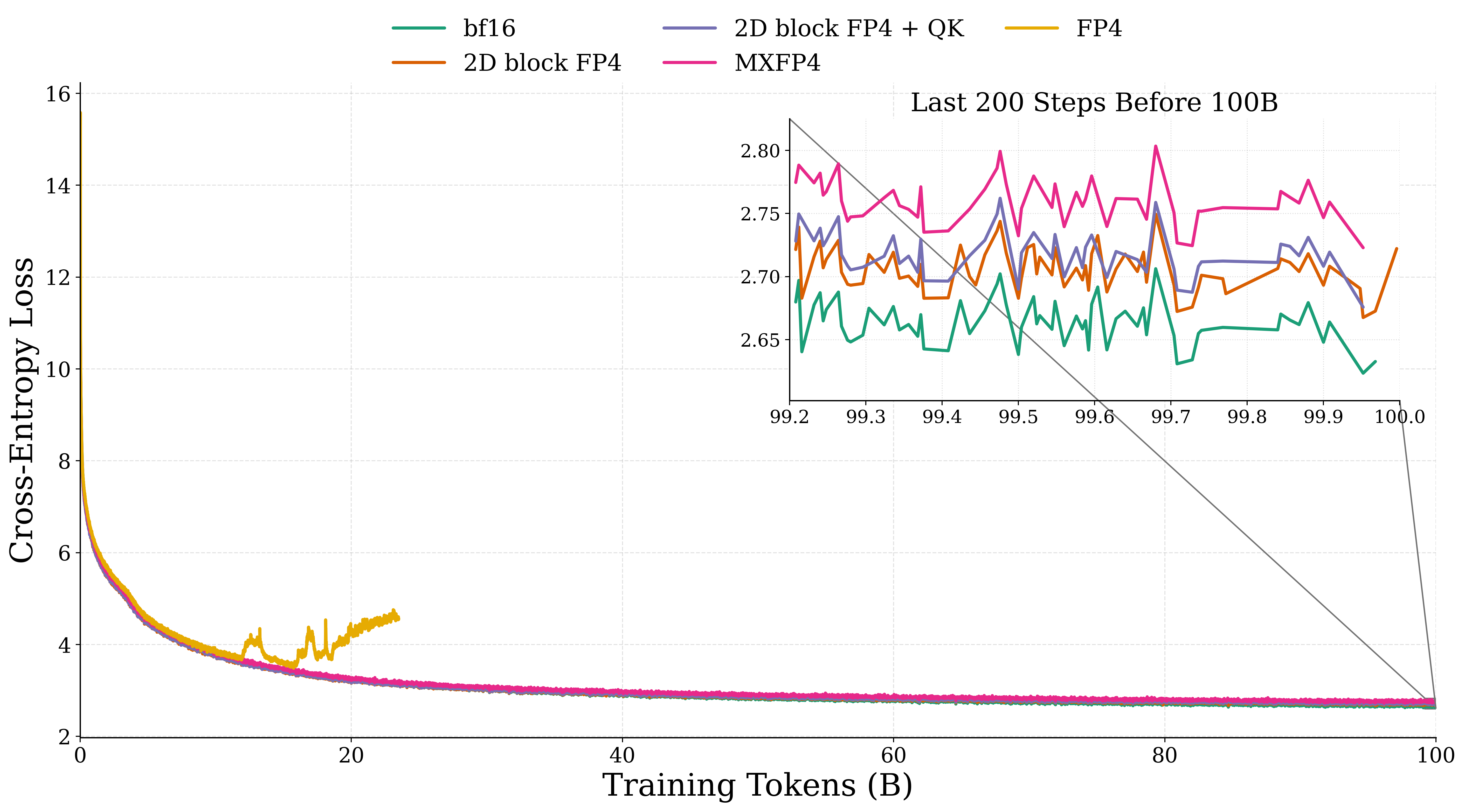}
        \caption{OLMo-1B}
        \label{fig:1b}
    \end{subfigure}
    
    \vspace{0.6em}
    
    \begin{subfigure}[t]{0.49\linewidth}
        \centering
        \includegraphics[width=\linewidth]{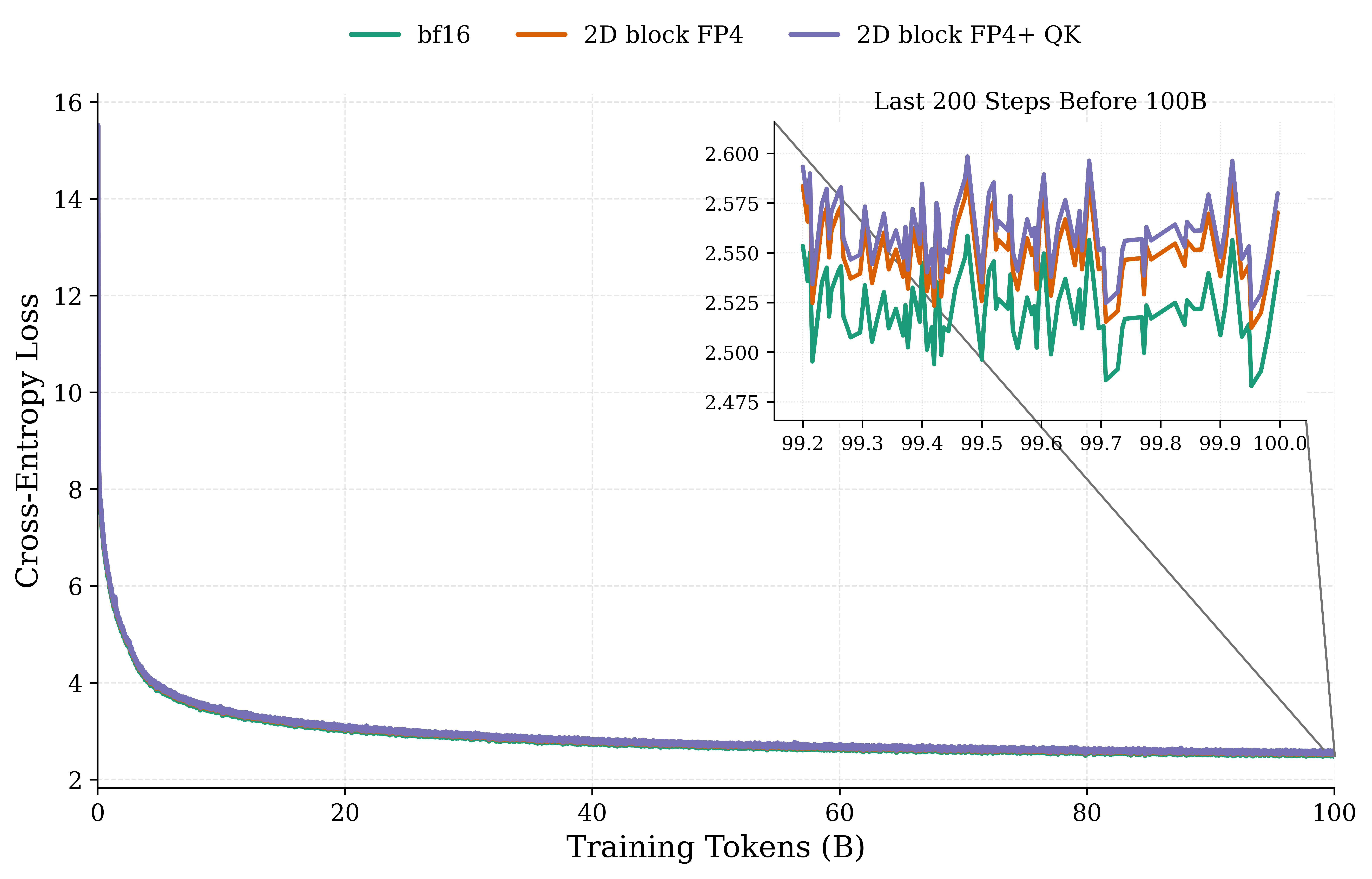}
        \caption{OLMo-7B}
        \label{fig:7b}
    \end{subfigure}
    \hfill
    \begin{subfigure}[t]{0.49\linewidth}
        \centering
        \includegraphics[width=\linewidth]{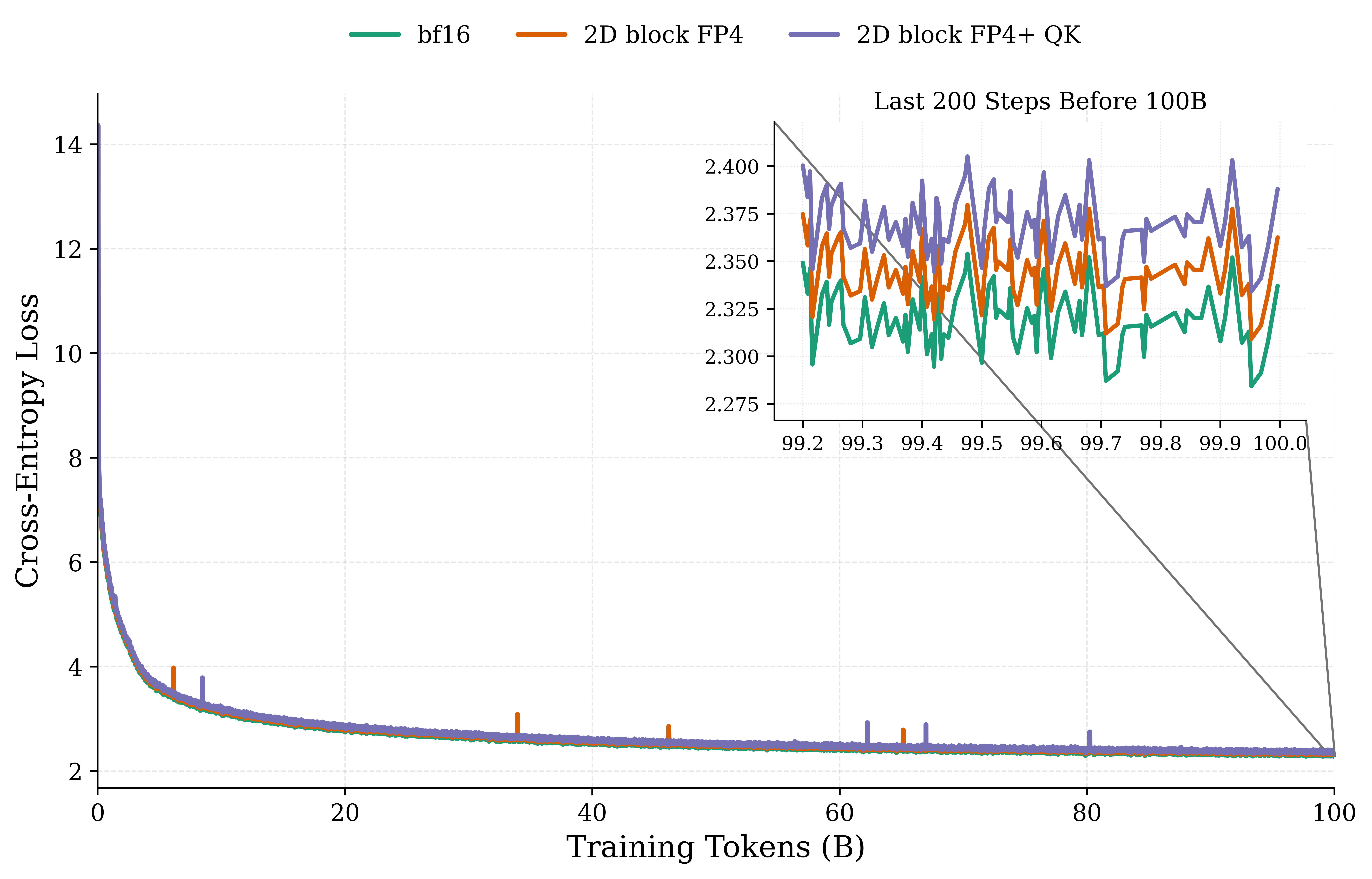}
        \caption{Qwen 30B MoE}
        \label{fig:30b}
    \end{subfigure}
    
    \caption{
    \textbf{Training dynamics under FP4 quantization across model scales.}
    Naive FP4 without microscaling diverges early, while microscaling-based methods remain stable. The proposed 2D block FP4 method consistently converges closest to BF16, with the performance gap decreasing as model size increases.
    }
    \label{fig:main_results}
\end{figure}

We begin by analyzing training stability and convergence. Figure~\ref{fig:main_results} shows the training dynamics across all model scales. In the OLMo-1B setting (Figure~\ref{fig:1b}), naive FP4 quantization without microscaling diverges early, with instability appearing at approximately 20B tokens. In contrast, all microscaling-based methods remain stable throughout training, confirming that block-wise scaling is necessary for FP4 optimization. Among these stable approaches, our 2D block FP4 method consistently tracks the BF16 baseline more closely than alternative microscaling schemes. At 1B scale, the final loss gap is approximately 1.1\%, compared to around 2\% for competing methods. Incorporating MXFP8 attention slightly increases the gap to 1.6\%, but remains substantially closer to BF16 than prior FP4 baselines. This suggests that enforcing transposition-invariant scaling is the primary factor driving stability, while higher-precision attention provides an additional but secondary benefit. This behavior remains consistent at larger scales. As shown in Figures~\ref{fig:7b} and \ref{fig:30b}, both OLMo-7B and Qwen 30B MoE converge stably under FP4 training, with final loss gaps of approximately 1\% relative to BF16. Notably, the gap decreases as model size increases, suggesting that larger models are more tolerant to quantization noise once scaling consistency is enforced.

\begin{figure}[t]
\centering
\includegraphics[width=0.55\linewidth]{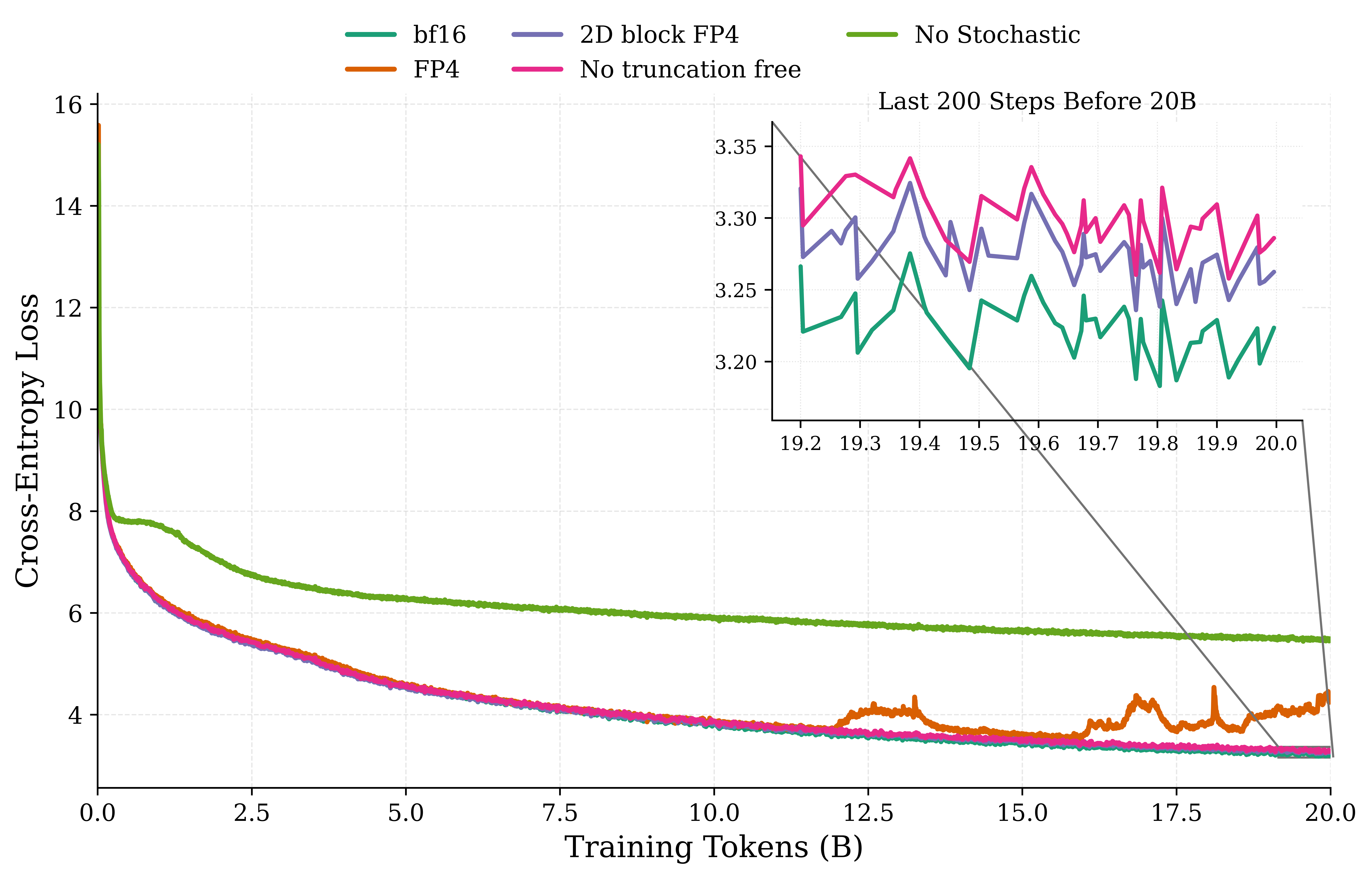}
\caption{
\textbf{Ablation of truncation-free scaling and stochastic rounding.}
Both components are necessary for stable FP4 training: removing truncation-free scaling leads to overflow-induced instability, while disabling stochastic rounding introduces bias and degrades convergence.
}
\label{fig:ablation_main}
\end{figure}

We next examine the contribution of truncation-free scaling and stochastic rounding. Figure~\ref{fig:ablation_main} shows that removing truncation-free scaling leads to instability due to numerical overflow, while disabling stochastic rounding introduces bias that degrades convergence. When both components are used together, the model achieves stable optimization and the smallest loss gap relative to BF16. These results highlight that stable FP4 training requires not only consistent scaling across forward and backward passes, but also careful control of overflow and rounding bias.

We then evaluate final model quality on language modeling benchmarks. Table~\ref{tab:lm_results} reports perplexity on Wikitext, Pile, and C4. Across all model scales, our method closely matches the BF16 baseline. For OLMo-1B, the average degradation is approximately 1.1\%, while for OLMo-7B and Qwen 30B the gap decreases to below 1\%. The hybrid 2D-FP4 + MXFP8 configuration exhibits slightly higher perplexity than the base 2D-FP4 variant, but remains within a narrow margin of BF16 across all datasets. Overall, these results show that aggressive precision reduction does not significantly degrade language modeling performance when scaling consistency is enforced.

\begin{table}[t]
\centering
\footnotesize
\caption{
\textbf{Language modeling results} (perplexity; lower is better).
}
\label{tab:lm_results}

\begin{tabular}{llccccc}
\toprule
Model & Method & $\Delta$ & Wikitext & Pile & C4 & Avg \\
\midrule
\multirow{3}{*}{OLMo-1B}
& BF16 & -- & 18.66 & 11.72 & 18.19 & 16.19 \\
& Ours & 1.1 & 18.82 & 11.80 & 18.57 & 16.40 \\
& +MXFP8 & 1.6 & 19.13 & 11.94 & 18.71 & 16.59 \\
\midrule
\multirow{3}{*}{OLMo-7B}
& BF16 & -- & 16.22 & 10.60 & 15.92 & 14.25 \\
& Ours & 0.9 & 16.30 & 10.66 & 16.05 & 14.34 \\
& +MXFP8 & 1.2 & 16.45 & 10.74 & 16.18 & 14.46 \\
\midrule
\multirow{3}{*}{Qwen 30B}
& BF16 & -- & 15.08 & 9.65 & 14.96 & 13.23 \\
& Ours & 0.8 & 15.15 & 9.70 & 15.05 & 13.30 \\
& +MXFP8 & 1.1 & 15.28 & 9.78 & 15.18 & 13.41 \\
\bottomrule
\end{tabular}
\end{table}

\begin{table}[t]
\centering
\footnotesize
\caption{
\textbf{Reasoning benchmark results} (accuracy; higher is better).
}
\label{tab:reasoning_results}

\begin{tabular}{llcccccc}
\toprule
Model & Method & SciQ & COPA & ARC-E & HellaSwag & Avg \\
\midrule
\multirow{3}{*}{OLMo-1B}
& BF16 & 77.60 & 70.00 & 51.75 & 45.52 & 61.22 \\
& Ours & 78.21 & 72.00 & 50.00 & 44.62 & 61.21 \\
& +MXFP8 & 77.22 & 69.00 & 49.64 & 44.15 & 60.00 \\
\midrule
\multirow{3}{*}{OLMo-7B}
& BF16 & 79.20 & 72.00 & 54.21 & 50.86 & 64.07 \\
& Ours & 79.80 & 73.00 & 53.80 & 50.30 & 64.23 \\
& +MXFP8 & 79.10 & 71.00 & 53.20 & 49.70 & 63.25 \\
\midrule
\multirow{3}{*}{Qwen 30B}
& BF16 & 85.40 & 79.31 & 60.79 & 56.96 & 70.62 \\
& Ours & 85.90 & 80.10 & 60.40 & 56.50 & 70.73 \\
& +MXFP8 & 85.20 & 78.90 & 59.80 & 55.90 & 69.95 \\
\bottomrule
\end{tabular}
\end{table}

We further evaluate downstream reasoning performance. Table~\ref{tab:reasoning_results} reports accuracy on SciQ, COPA, ARC-Easy, and HellaSwag. Across all tasks, our method remains close to BF16, with performance differences generally within approximately 1 percentage point. In several cases, particularly on SciQ and COPA, the FP4 model slightly exceeds the BF16 baseline. While modest, this improvement may be attributed to a regularization effect induced by quantization noise. The MXFP8 variant exhibits slightly lower accuracy than the base FP4 configuration, but remains competitive overall. We also analyze the trade-off between block size, accuracy, and efficiency. Table~\ref{tab:block_tradeoff} summarizes  trends for different 2D block sizes. Smaller blocks improve numerical fidelity but increase metadata overhead and reduce throughput, while larger blocks improve efficiency at the cost of increased loss gap. The $32\times32$ configuration used in our experiments provides a practical balance between stability and efficiency. Taken together, these results demonstrate that enforcing forward--backward consistent scaling enables stable end-to-end FP4 training while preserving model quality across both dense and MoE architectures. Once scale inconsistency is removed, FP4 training can closely approach BF16 performance.

\begin{table}[t]
\centering
\footnotesize
\caption{
\textbf{Block-size trade-offs in 2D FP4 quantization.}
}
\label{tab:block_tradeoff}

\begin{tabular}{lccc}
\toprule
Block size & Final loss gap (\%) & Relative memory & Relative throughput \\
\midrule
$8 \times 8$   & $\sim$0.8--1.0 & $\sim$0.30$\times$ & $\sim$0.85$\times$ \\
$16 \times 16$ & $\sim$0.9--1.1 & $\sim$0.27$\times$ & $\sim$0.92$\times$ \\
$32 \times 32$ & 1.1            & $\sim$0.25$\times$ & $1.00\times$ \\
$64 \times 64$ & $\sim$1.3--1.6 & $\sim$0.24$\times$ & $\sim$1.05$\times$ \\
\bottomrule
\end{tabular}
\end{table}

\section{Discussion and Limitations}

Our results suggest that the main challenge in FP4 training is not simply limited numerical precision, but a structural mismatch between quantization schemes and backpropagation. By enforcing transposition-invariant scaling through 2D block quantization, we eliminate a key source of gradient bias and enable stable optimization at 4-bit precision. This perspective reframes low-precision training as a problem of \emph{consistency}, rather than purely representational fidelity.  The proposed approach is simple and broadly applicable, requiring minimal modifications to existing training pipelines while delivering consistent improvements across model scales and architectures. Notably, the performance gap between FP4 and higher-precision training decreases as model size increases, suggesting that larger models are inherently more robust to quantization noise once systematic bias is removed. Furthermore, our mixed-precision treatment of attention demonstrates that different components of the transformer exhibit varying sensitivity to aggressive quantization, indicating that selective precision allocation is an effective and practical strategy.

Despite these encouraging findings, several limitations remain. First, our experiments rely on software emulation of FP4 behavior on hardware that does not natively support the MXFP4 format. Real-world efficiency gains will require next-generation accelerators with native support, where microscaling formats can be executed efficiently to fully realize the speed and energy benefits of FP4 training.  Second, while 2D block quantization improves consistency in matrix operations, it introduces additional design trade-offs, including the choice of block size and the overhead of storing scaling factors. The optimal configuration is likely to depend on the specific architecture and workload, and may require further tuning in practice.  Finally, our evaluation is limited to models up to 30B parameters and training runs of up to 100B tokens. Scaling to larger, frontier-scale models and longer training regimes may expose additional stability challenges. Future work should explore adaptive or learned block structures, closer hardware–software co-design, and the possibility of further reducing precision in sensitive components such as attention.

\section{Conclusion}

We presented a low-precision training framework for large language models based on transposition-invariant 2D block FP4 quantization, combined with MXFP8 for attention. Our central insight is that instability in existing FP4 methods arises not only from limited precision, but from scale inconsistency between forward and backward passes induced by tensor transposition. By replacing 1D microscaling with 2D square blocks, our method enforces consistent scaling and substantially improves optimization stability. Across dense models up to 7B parameters and a 30B Mixture-of-Experts model trained on up to 100B tokens, our approach enables stable end-to-end FP4 training while closely matching BF16 performance, with less than 1.3\% degradation. Ablations further show that truncation-free scaling and stochastic rounding are important complementary components for achieving robust convergence under extreme quantization. More broadly, our results suggest that the main obstacle to practical FP4 training is not simply quantization error, but inconsistency in scaling across forward and backward computations. This perspective shifts the design of low-precision training from improving local representational fidelity to enforcing global computational consistency. We believe this view opens promising directions for future work, including hardware support for transposition-consistent quantization and further scaling to larger models.

\bibliographystyle{plain}
\bibliography{references}

\newpage
\appendix

\section{Supplementary Material}
\label{app:supplementary}

This appendix provides additional implementation and evaluation details for the proposed transposition-invariant FP4 training method. We focus on the details most relevant for reproducibility: training configuration, quantization placement, scaling and rounding rules, evaluation protocol, ablation interpretation, and the scope of the software-emulation results.

\section{Training Configuration}
\label{app:training_configuration}

Table~\ref{tab:training_configurations} summarizes the training setup used for all model scales. All models are trained with a sequence length of 2K tokens and a token budget of 100B. For each model, the BF16, 2D-FP4, and 2D-FP4+MXFP8 variants use the same optimizer, learning-rate schedule, warmup, and global batch size so that differences in training dynamics are attributable to the numerical format rather than to optimization hyperparameters.

\begin{table}[h]
\centering
\caption{Training configurations for all experiments.}
\label{tab:training_configurations}
\resizebox{\textwidth}{!}{
\begin{tabular}{lccc}
\toprule
\textbf{Configuration} & \textbf{OLMo-1B} & \textbf{OLMo-7B} & \textbf{Qwen 30B MoE} \\
\midrule
\multicolumn{4}{l}{\textbf{Model \& Data}} \\
Training Tokens & 100B & 100B & 100B \\
Sequence Length & 2K & 2K & 2K \\
Objective & Causal LM & Causal LM & Causal LM \\
\midrule
\multicolumn{4}{l}{\textbf{Optimization}} \\
Optimizer & AdamW & AdamW & AdamW \\
Peak Learning Rate & $4.0 \times 10^{-4}$ & $3.0 \times 10^{-4}$ & $1.0 \times 10^{-4}$ \\
Final Learning Rate & $4.0 \times 10^{-5}$ & $3.0 \times 10^{-5}$ & $1.0 \times 10^{-5}$ \\
LR Schedule & Cosine Decay & Cosine Decay & Cosine Decay \\
Warmup Steps & 4K & 4K & 4K \\
Global Batch Size & 2048 & 2048 & 512 \\
Weight Decay & 0.02 & 0.02 & 0.02 \\
Gradient Clipping & 1.0 & 1.0 & 1.0 \\
Optimizer State Precision & BF16 & BF16 & BF16 \\
\midrule
\multicolumn{4}{l}{\textbf{Quantization}} \\
Weight Quantization & 2D-FP4 & 2D-FP4 & 2D-FP4 \\
Gradient Quantization & 2D-FP4 & 2D-FP4 & 2D-FP4 \\
Activation Quantization & 1D-FP4 & 1D-FP4 & 1D-FP4 \\
Query/Key Precision & MXFP8 & MXFP8 & MXFP8 \\
Default 2D Block Size & $32 \times 32$ & $32 \times 32$ & $32 \times 32$ \\
Backward Rounding & Stochastic & Stochastic & Stochastic \\
Scaling Rule & Truncation-free & Truncation-free & Truncation-free \\
\bottomrule
\end{tabular}
}
\end{table}

\section{Quantization Placement}
\label{app:quantization_placement}

\paragraph{Linear layers.}
For a linear layer,
\[
Y = XW^\top,
\qquad
\nabla X = \nabla YW,
\qquad
\nabla W = (\nabla Y)^\top X.
\]
These computations involve transposed tensor views during backpropagation. We therefore apply 2D block quantization to weights and gradient tensors so that the same values retain consistent scale assignments across forward and backward matrix multiplications.

\paragraph{Activations.}
Activations use a 1D block layout. This avoids grouping semantically unrelated hidden channels into square 2D blocks. In practice, applying 2D blocks to weights and gradients provides the main forward--backward consistency benefit, while 1D activation quantization preserves a simpler channel-aligned layout.

\paragraph{Attention.}
The query and key projections use MXFP8 because small perturbations in $Q$ and $K$ can be amplified by $QK^\top$ and then by the softmax. The value projection, output projection, and MLP projections use 2D-FP4.

\section{2D Block Scaling and Transposition Consistency}
\label{app:transposition_consistency}

Let $X \in \mathbb{R}^{m \times n}$ be partitioned into square blocks of size $b \times b$. For block indices $(i,j)$, define
\[
B_{i,j}
=
X[ib:(i+1)b,\; jb:(j+1)b].
\]
The scale for a block is computed from the maximum absolute value in that block:
\[
M_{i,j} = \max_{x \in B_{i,j}} |x|.
\]
Using truncation-free scaling, the block scale is
\[
S_{i,j}
=
2^{\left\lceil
\log_2
\left(
\frac{2M_{i,j}}{Q_p - Q_n}
\right)
\right\rceil},
\]
where $Q_p$ and $Q_n$ are the positive and negative representable FP4 bounds. The quantized block is
\[
\widehat{B}_{i,j}
=
S_{i,j}
\cdot
\mathrm{round}_{\mathrm{FP4}}
\left(
\frac{B_{i,j}}{S_{i,j}}
\right).
\]

Under transposition, $B_{i,j}$ maps to a block at the swapped block index:
\[
B_{i,j}^{\top}
=
X^\top[jb:(j+1)b,\; ib:(i+1)b].
\]
The transposed block contains the same values as the original block, so its maximum absolute value is unchanged:
\[
M(B_{i,j}) = M(B_{i,j}^{\top}).
\]
Therefore, the scale is also unchanged:
\[
S(B_{i,j}) = S(B_{i,j}^{\top}).
\]
This is the transposition-invariance property used by our method. In contrast, a 1D block layout generally changes which values are grouped together after transposition, causing the same value to be quantized with different scales in the forward and backward passes.

\paragraph{Boundary blocks.}
If a tensor dimension is not divisible by $b$, the boundary block contains fewer than $b \times b$ valid entries. The scale is computed only over valid entries. Padding values, when used for implementation convenience, are excluded from the scale computation.

\section{Rounding and Gradient Propagation}
\label{app:rounding_gradient}

During the forward pass, scaled values are rounded to the nearest representable FP4 value. During the backward pass, we use stochastic rounding for gradient tensors. For a value $x$ between adjacent representable values $a$ and $b$, stochastic rounding chooses $a$ or $b$ with probabilities that preserve the expectation:
\[
\mathbb{E}[\mathrm{round}_{\mathrm{stoch}}(x)] = x.
\]
This reduces systematic rounding bias in gradient updates. Gradients are propagated through quantization using the straight-through estimator:
\[
\frac{\partial \mathcal{L}}{\partial X}
\approx
\frac{\partial \mathcal{L}}{\partial \widehat{X}}.
\]
Thus, quantization is applied in the numerical computation, while the backward derivative of the quantizer is approximated as the identity.

\section{Evaluation Protocol}
\label{app:evaluation_protocol}

\paragraph{Language modeling.}
We evaluate language modeling quality using perplexity on Wikitext, Pile, and C4. Perplexity is computed as
\[
\mathrm{PPL}
=
\exp
\left(
-\frac{1}{N}
\sum_{t=1}^{N}
\log p(x_t \mid x_{<t})
\right),
\]
where $N$ is the number of evaluated tokens. Lower perplexity indicates better language modeling performance. Within each model family, BF16 and low-precision variants use the same tokenizer, sequence length, and evaluation batches.

\paragraph{Reasoning tasks.}
We evaluate downstream reasoning using SciQ, COPA, ARC-Easy, and HellaSwag. Accuracy is computed as
\[
\mathrm{Accuracy}
=
\frac{\text{number of correct predictions}}
{\text{number of evaluated examples}}.
\]
For multiple-choice benchmarks, the predicted answer is selected using normalized likelihood over candidate answers. The same scoring procedure is used for BF16, 2D-FP4, and 2D-FP4+MXFP8.

\paragraph{Relative degradation.}
For perplexity, the relative degradation from BF16 is
\[
\Delta_{\mathrm{PPL}}(\%)
=
100
\times
\frac{
\mathrm{PPL}_{\mathrm{method}}
-
\mathrm{PPL}_{\mathrm{BF16}}
}{
\mathrm{PPL}_{\mathrm{BF16}}
}.
\]
For accuracy, we report the absolute difference from BF16:
\[
\Delta_{\mathrm{Acc}}
=
\mathrm{Acc}_{\mathrm{method}}
-
\mathrm{Acc}_{\mathrm{BF16}}.
\]

\section{Ablation Interpretation}
\label{app:ablation_interpretation}

\paragraph{Block size.}
The default block size is $32 \times 32$. Smaller blocks, such as $8 \times 8$ and $16 \times 16$, reduce quantization error because each scale covers fewer values, but they increase scale metadata and can reduce throughput. Larger blocks, such as $64 \times 64$, reduce metadata overhead but increase the final loss gap. The $32 \times 32$ setting provides a practical balance between stability, memory overhead, and matrix-multiplication tiling.

\paragraph{Truncation-free scaling.}
Removing truncation-free scaling makes FP4 training more vulnerable to clipping and overflow. Since FP4 has limited dynamic range, even a small number of clipped high-magnitude values can destabilize training. The ablation in the main paper shows that truncation-free scaling is necessary in addition to the 2D block structure.

\paragraph{Stochastic rounding.}
Replacing stochastic rounding with deterministic rounding increases bias in gradient tensors. This leads to worse convergence and a larger gap from BF16. The ablation shows that stable FP4 training requires both scale consistency and approximately unbiased gradient quantization.

\paragraph{MXFP8 attention.}
MXFP8 attention is included to improve numerical robustness in the query/key pathway. The main results show that 2D-FP4 alone is already very close to BF16, while the MXFP8 variant can slightly increase the final perplexity gap in some settings. We therefore view MXFP8 attention as a stability-oriented mixed-precision option, while the primary contribution remains transposition-invariant 2D-FP4 scaling.

\end{document}